\documentclass{article}
\usepackage{spconf,amsmath,graphicx}

\usepackage{subfigure} 
\usepackage{amsmath,amsthm}
\usepackage{slashbox}
\usepackage{multirow}
\usepackage{amssymb}

\newcommand{\argmax}{\operatornamewithlimits{argmax}}

\usepackage{hyperref}

\title{Word Recognition with Deep Conditional Random Fields}
\name{Gang Chen$^{\dagger}$ \qquad Yawei Li$^{\star}$ \qquad Sargur N. Srihari$^{\dagger}$}

\address{$^{\dagger}$ Department of Computer Science, SUNY at Buffalo, Buffalo NY 14260\\
    $^{\star}$ School of Communication and Information Engineering \\
    University of Electronic Science and Technology of China \\Chengdu, Sichuan 611731 China}
\begin{document}
%
\title{Word Recognition with Deep Conditional Random Fields }


%


\maketitle

\begin{abstract}
Recognition of handwritten words continues to be an important problem in document analysis and recognition. Existing approaches extract hand-engineered features from word images--which can perform poorly with new data sets. Recently, deep learning has attracted great attention because of the ability to learn features from raw data. Moreover they have yielded state-of-the-art results in classification tasks including character recognition and scene recognition. On the other hand, word recognition is a sequential problem where we need to model the correlation between characters. In this paper, we propose using deep Conditional Random Fields (deep CRFs) for word recognition. Basically, we combine CRFs with deep learning, in which deep features are learned and sequences are labeled in a unified framework. We pre-train the deep structure with stacked restricted Boltzmann machines (RBMs) for feature learning and optimize the entire network with an online learning algorithm. The proposed model was evaluated on two datasets, and seen to perform  significantly better than competitive baseline models.
\end{abstract}


\section{Introduction}
Word recognition \cite{LeCun89,Plamondon00,Shetty07} can be formulated as a sequence labeling problem. Each word image is a set of candidate character segments obtained first by segmentation methods, then it is labeled with classifiers, such as support vector machines (SVM) and Conditional Random Fields (CRFs). Although it has been researched many years, it is still an challenge problem, considering the complex writing styles. For example, the poor quality in handwritten documents makes character boundaries hard to determine \cite{LeCun98}. What's more, even given the segmented characters in the word images, it is still hard to get satisfied results, because different people have different handwriting styles. In this paper, we attempt to address the second issue. In other words, we assume character segmentation is given, and then directly recognize the entire word without character segmentation, as \cite{Shetty07} did.

Recently, deep learning has attracted great attention because it can learn features automatically from raw data, which has been thought as a vital step forward to artificial intelligence (AI). 
Linear CRFs, as a graphic model with correlation among labels, has been a powerful tool for sequential labeling, with applications on a wide range of tasks, such as character recognition, speech recognition and natural language processing. Moreover, as a discriminative model, it has shown advantages over generative models, i.e. HMMs on labeling sequence data \cite{Feng06,Maaten11}. Thus, we can leverage CRFs for word recognition, because it has shown promising results on a variety of handwriting datasets \cite{Feng06,Shetty07}. 

Inspired by deep learning for feature learning, we unify deep learning and CRFs into one framework, so that it can discover discriminative features to improve word classification task. 
Thus, our model is more powerful than linear Conditional Random Fields (CRFs) because the objective function learns latent non-linear features so that target labeling can be better predicted. Different from traditional approaches, we train our model with an independent learning stage and also use online learning to estimate model parameters. We test our method on two handwriting datasets, and show the advantages of our method over shallow CRFs models and deep learning methods significantly. 

\section{Related work}
Handwriting recognition \cite{Plamondon00} is a classical recognition problem, which has been researched for a long time. However, considering the complex cases, such as cursive styles and spurious strokes, it is far from being solved. 

Over the past decades, many methods have beed proposed \cite{LeCun89,Hinton06b} for handwriting recognition, and good success has been achieved for small-vocabulary and highly constrained domains such as digital recognition \cite{Hinton06b}, mail sorting \cite{LeCun89,Plamondon00} and check processing \cite{Madhvanath01}. Marti et al proposed to use Hidden Markov model (HMM) to incorporate the context information for handwritten material recognition \cite{Marti01}. In their method, each character has 14 states, and words are modeled as a concatenation of these states under HMM. Later, Vinciarelli et al proposed a sliding window approach \cite{Vinciarelli03}, an high order n-gram HMM model (up to trigrams) and demonstrated promising results on large vocabulary handwritten data. Boosted decision trees \cite{Howe05} had been used on word recognition and retrieval, and yielded good retrieval performance with low word error rate. Inspired by the advantages of CRFs \cite{Lafferty01} for sequential labeling problem, several methods were put forward recently. For example, \cite{Feng06} leveraged CRFs for handwriting recognition, and demonstrated that on the whole word recognition task, linear CRFs performs better than HMM. Similarly, another CRFs with dynamical programming \cite{Shetty07} was also proposed to word recognition. 

More recently, deep learning methods, such as CNN, DBN and recurrent neural network (RNN) \cite{Hochreiter97} had gained significant attention on classification tasks. For example, Graves et al \cite{Graves08} used the bi-directional and multi-dimensional long short term memory (LSTM) \cite{Karpathy15} and yielded promising results on Arabic handwriting recognition. Furthermore, Hidden conditional random fields \cite{Maaten11} was proposed and yielded the state of the art results on the upenn OCR dataset \cite{Taskar03}. CRFs also has been combined with neural network for sequential labeling (DNN+CRFs \cite{Do10}).

\section{Word Recognition with Deep CRFs}
Let $D = \{ \langle {\bf x}_{i}, {\bf y}_{i}\rangle \}_{i=1}^N$ be a set of $N$ training examples. Each example is a pair of a time series $\langle {\bf x}_{i}, {\bf y}_{i}\rangle$, with ${\bf x}_{i} = \{{\bf x}_{i,1}, {\bf x}_{i,2},...,{\bf x}_{i,T_i}\}$ and ${\bf y}_{i} = \{{ y}_{i,1},{y}_{i,2},...,{y}_{i,T_i}\}$, where ${\bf x}_{i,t}  \in \mathbb{R}^d$ is the $i$-th observation at time $t$ and ${y}_{i,t}$ is the corresponding label (we indicate its encoded vector as ${\bf y}_{i,t}$ that uses a so-called 1-of-$K$ encoding).

\subsection{Linear CRFs}
Linear first-order CRFs \cite{Lafferty01} is a conditional discriminative model over the label sequence given the data 
\begin{equation}
\centering
p({\bf y}_{i} | {\bf x}_{i}) = \frac{\textrm{exp}\{E({\bf x}_{i}, {\bf y}_{i}) \}}{Z({\bf x}_{i})}
\label{eq:eqcrf}
\end{equation}
where $Z({\bf x}_{i})$ is the partition function and $E( {\bf x}_{i}, {\bf y}_{i})$ is the energy function given by
\begin{align}
&E( {\bf x}_{i}, {\bf y}_{i}) =  {\bf y}_{i,1}^T \boldsymbol{\pi} + {\bf y}_{i, T_i}^T \boldsymbol{\tau} \nonumber \\
+ & \sum_{t=1}^{T_i}  ({\bf x}_{i,t}^T {\bf W} {\bf y}_{i,t}  + {\bf b}^T {\bf y}_{i,t} )  + \sum_{t=2}^{T_i}  {\bf y}_{i, t-1}^T {\bf A} {\bf y}_{i,t}
\label{eq:eqe}
\end{align}
where ${\bf y}_{i,1}^T \boldsymbol{\pi}$ and ${\bf y}_{i, T_i}^T \boldsymbol{\tau}$ are the initial-state and final-state factors respectively, ${\bf b}^T {\bf y}_{i,t}$ is the bias term for labels, ${\bf A} \in \mathbb{R}^{K \times K}$ represents the state transition parameters and ${\bf W} \in \mathbb{R}^{d \times K}$ represents the classification parameter of the data-dependent term.
One of the main disadvantages of linear CRFs is the linear dependence on the raw input data term. Thus, we introduce our sequential labeling model with deep feature learning, which leverages both context information, as well as the nonlinear representations in the deep learning architecture \cite{Hinton06b}. 

\subsection{Deep CRFs}
Although it is possible to leverage the deep neural networks for structured prediction, its output space is explosively growing because of non-determined length of sequential data. Thus, we consider a compromised model, which combines CRFs and deep learning in an unified framework. 
Thus, we propose an objective function with $L$ layers neural network structure, 
\begin{align}
 \mathcal{L}(D; \boldsymbol{\theta}, \boldsymbol{\omega}) &= - \sum_{i=1}^N \textrm{log} p({\bf y}_{i,1},..., {\bf y}_{i,T_i}| {\bf h}_{i,1},...,  {\bf h}_{i,T_i})  \nonumber \\
 &+  \lambda_2 ||\boldsymbol{ \theta}||^2 + \lambda_3 ||\boldsymbol{\omega}||
\label{eq:eqobj}
\end{align}
where $\boldsymbol{\theta}$ and $\boldsymbol{\omega}$ are the top layer parameters and lower layer ($l = \{1,...,L-1\}$) parameters respectively, which will be explained later. The first row on the right side of the equation is from the linear CRFs  in Eq. (\ref{eq:eqcrf}), but with latent features, which depends respectively on $\boldsymbol{\theta}$ and the latent non-linear features ${\bf h}_i = \{{\bf h}_{i,1},..,{\bf h}_{i,T_i} \}$ in the coding space, with
\begin{align}
& \textrm{log} p({\bf y}_{i,1},..., {\bf y}_{i,T_i}| {\bf h}_{i,1},...,  {\bf h}_{i,T_i})   \nonumber \\
= & \sum_{t=2}^{T_i}  {\bf y}_{i, t-1}^T {\bf A} {\bf y}_{i,t} +   \sum_{t=1}^{T_i}  \big({\bf h}_{i,t}^T {\bf W} {\bf y}_{i,t}  + {\bf b}^T {\bf y}_{i,t}  \big)  \nonumber \\
+ & {\bf y}_{i,1}^T \boldsymbol{\pi} + {\bf y}_{i,T_i}^T \boldsymbol{\tau} - \textrm{log} (Z({\bf h}_{i}))
\label{eq:linearpart}
\end{align}
and non-linear mappings ${\bf h}_{i}$ is the output with $L-1$ layers neural network, s.t. 
\begin{equation}\label{eq:hidden}
{\bf h}_{i} = \underbrace{f_{L-1} \circ f_{L-2}\circ \cdot\cdot\cdot \circ f_{1}}_{L-1 \textrm{ times}}({\bf x}_{i})
\end{equation}
where $\circ$ indicates the function composition, and $f_i$ is logistic function with the weight parameter ${\bf W}_l$ respectively for $l = \{1,..,L-1\}$, refer more details in Sec. \ref{sec:learning}. With a bit abuse of notation, we denote ${\bf h}_{i,t}  = f_{1 \rightarrow (L-1)} ({\bf x}_{i,t})$.

The last two terms in Eq. \ref{eq:eqobj} are for regularization on the all parameters
with $\boldsymbol{\theta} = \{ {\bf A}, {\bf W}, \boldsymbol{\pi}, \boldsymbol{\tau}, \boldsymbol{b}, \boldsymbol{c}\}$, and $\boldsymbol{\omega} = \{ {\bf W}_l |  l \in [1,..,L-1] \} $. We add the $\ell_2$ regularization to $\boldsymbol{\theta}$ as most linear CRFs does, while we have the $\ell_1$-regularized term on weight parameters $\boldsymbol{\omega}$ in the deep neural network to avoid overfitting in the learning process.

The aim of our objective function in Eq. (\ref{eq:eqobj}) is for sequential labeling, which explores both the advantages of Markov properties in CRFs and latent representations in deep learning. Firstly, our model can learn non-linear representation and label sequences with non-determined length. Secondly, our model can predict structured outputs or label sequences, while the DBN model \cite{Hinton06b} is just one label for each instance, which is independent without context information. Note that we use the first-order CRFs for clarity in Eq. \ref{eq:linearpart} and the rest of the paper, which can be easily extended for the second or high-order cases. Lastly, compared to traditional DNN+CRFs \cite{Do10}, we use an online algorithm in our deep learning model for parameter updating, which has the potential to handle large scale dataset.

\subsection{Learning}\label{sec:learning}
We take an online learning strategy in our method, which is different from traditional approaches. We use RBMs to initialize the weights for $l = \{1,..,L-1\}$ layer by layer greedily, with contrast divergence \cite{Hinton06a} (we used CD-1 in our experiments). Then we compute the sub-gradient w.r.t. $\boldsymbol{\theta}$ and $\boldsymbol{\omega}$ in the objective function, and optimize it with online learning. 

{\bf Initialization}: In our deep model, the weights from the layers $1$ to $L-1$ are ${\bf W}_l$ respectively, for $l = \{1,..,L-1\}$, and the top layer $L$ has weight ${\bf W}$. We first pre-train the $L$-layer deep structure with RBMs layer by layer greedily. 

{\bf Learning}:
In training the CRFs with deep feature learning, our aim is to minimize objective function $\mathcal{L}(D; \boldsymbol{\theta}, \boldsymbol{\omega}) $ in Eq. (\ref{eq:eqobj}). Because we introduce the deep neural network here for feature learning, the objective is not convex function anymore. However, we can find a local minimum in Eq. (\ref{eq:eqobj}). In our learning framework, we optimize the objective function with an online learn algorithm, by mixing perceptron training and stochastic gradient descent. 

Firstly, we can calculate the (sub)gradients for all parameters. Considering different regularization methods for $\boldsymbol{\theta}$ and $\boldsymbol{\omega}$ respectively, we can calculate gradients w.r.t. them separately. As for the parameters in the negative log likelihood in Eq. \ref{eq:eqobj}, we can compute the gradients w.r.t. $\boldsymbol{\theta}$ as follows
\begin{subequations}
\begin{equation}
\frac{\mathcal{\partial L}}{\partial {\bf A}} = \sum_{i=1}^N \sum_{t=2}^{T_i} {\bf y}_{i,t-1}  ({\bf y}_{i,t})^T - \boldsymbol{\gamma}_{i,t-1}  (\boldsymbol{\gamma}_{i,t})^T;  \label{eq:grad0}
\end{equation}
\begin{align}
\frac{\mathcal{\partial L}}{\partial {\bf W}} & = \sum_{i=1}^N \sum_{t =1}^{T_i}  \big( {\bf h}_{i,t} ( {\bf y}_{i,t} - \boldsymbol{\gamma}_{i,t})^T 
 \end{align}
\label{eq:grad}
\end{subequations} 
where ${\bf y}_{i,t}  \in \mathbb{R}^K$ is the 1-of-$K$ encoding for labeling, $\boldsymbol{\gamma}_{i,t} \in \mathbb{R}^K$ is the vector of length $K$, which is the posterior probability for labels in the sequence and will be introduced in Sec. \ref{sec:infer}; and ${\bf h}_{i} = \{ {\bf h}_{i,1},...,{\bf h}_{i,T_i} \}$ are the latent features learned via Eq. (\ref{eq:hidden}). In the above gradients, we have ignored the gradients w.r.t. biases for convenience. Note that it is easy to derive the gradients of the $\ell_2$ regularization term w.r.t. $\boldsymbol{\theta}$ in the objective in Eq. (\ref{eq:eqobj}), which can be added to the gradients in Eq. (\ref{eq:grad}). 

As for the gradients of weights $\boldsymbol{\omega} = \{ {\bf W}_l |  l \in [1,..,L-1] \} $, we first use backpropagation to get the partial gradient in the neural network, refer to \cite{Hinton06b} for more details. Then the gradient of the $\ell_1$ term in Eq. (\ref{eq:eqobj}) can be attached to get the final gradients w.r.t. ${\bf W}_{l}$ for $l=\{1,..,L-1\}$.

Finally, we use a mixture of perceptron learning and stochastic gradient descent to optimize the objective function. 
In our experiments, we tried L-BFGS, but it can be easily trapped into the bad local minimum, and performs worse than other optimization methods in almost all experiments. Thus, in this work, we use perceptron-based learning for the CRF related parameters and stochastic gradient descent for the parameters in the deep structure in all our experiments. 

Thus, for the CRF related parameters $\boldsymbol{\theta}$ in Eq. (\ref{eq:eqobj}), we first project ${\bf x}_i$ into the code ${\bf h}_{i}$ according to Eq. (\ref{eq:hidden}). Then, the updating rule takes the form below
\begin{equation}
\boldsymbol{\theta} \leftarrow  \boldsymbol{\theta}  + \eta_{\boldsymbol{\theta} }   \frac{\partial }{\partial \boldsymbol {\theta}} \big({E({\bf h}_{i}, {\bf y}_{i}) - E({\bf h}_{i}, {\bf y}_i^{\ast})} \big)
\label{eq:theta}
\end{equation}
where ${\bf y}_i^{\ast}$ is the most violated constraint in the misclassificated case, and $\eta_{\boldsymbol{\theta} }$ is a parameter step size. Note that the posterior probability $\boldsymbol{\gamma}_{i,t} \in \mathbb{R}^K$ in Eq. (\ref{eq:grad}) should be changed into the hard label assignment ${\bf y}_{i,t}^{\ast}$ in the inference stage. 

While for the weights $\boldsymbol{\omega}$ in the deep neural network, we first use backpropagation to compute the gradients, and then update it as follows
\begin{equation}
\boldsymbol{\omega} \leftarrow  \boldsymbol{\omega}  - \eta_{\boldsymbol{\omega} }   \frac{\partial \mathcal{L}}{\partial \boldsymbol {\omega}}
\label{eq:omega}
\end{equation}
where $\eta_{\boldsymbol{\omega} }$ is the step size for the parameters.

\subsection{Inference}\label{sec:infer}
In the testing stage, the main inferential problem is to compute the most likely label sequence ${\bf y}_{1,...,T}^{\ast}$ given the data ${\bf x}_{1,...,T}$ by $\argmax_{{{\bf y}\prime}_{1,...,T}} p({{\bf y}\prime}_{1,...,T} | {\bf x}_{1,...,T} )$.

Given any new sequence ${\bf x}_{i} = \{ {\bf x}_{i,1},...,{\bf x}_{i,T_i} \}$, we first use Eq. (\ref{eq:hidden}) to compute the non-linear code ${\bf h}_{i} = \{ {\bf h}_{i,1},...,{\bf h}_{i,T_i} \}$. Then, we can do the inference as linear CRFs does by thinking ${\bf h}_{i} $ as the new obersevation. The inference problem can be
formulated as 
\begin{equation}
{\bf y}_{1,...,T}^{\ast} = \argmax_{{{\bf y}\prime}_{1,...,T_i}} p({{\bf y}\prime}_{1,...,T_i} | {\bf h}_{i} )
\end{equation}
This can be solved efficiently with Viterbi algorithm \cite{Rabiner89,Bishop06}. Note that $\boldsymbol{\gamma}_{i,t} = p( {\bf y}_{i,t}^{\ast} | {\bf h}_{i} ) $ is the posterior probability from Viterbi algorithm, and can be used in Eq. \ref{eq:grad} to calculate the gradient w.r.t. $\boldsymbol{\theta}$.

\section{Experiments} 
To test our method, we compared our method to the state of the art approaches and performed experiments on word recognition task on two widely used datasets: OCR dataset and ICDAR2003 word recognition dataset. 

\subsection{Data sets}
1. The OCR dataset \cite{Taskar03} contains data for 6, 877 handwritten words with 55 unique words, in which each word ${\bf x}_i$ is represented as a series of handwritten characters $\{ {\bf x}_{i1},...,{\bf x}_{i, T_i} \}$. The data consists of a total of 52, 152 characters (i.e., frames), with 26 unique classes. Each character is a binary image of size $16 \times 8$ pixels, leading to a 128-dimensional binary feature vector. 
In our experiments, we used the four data sets in \cite{Maaten11}, which are available on the author's website\footnote{\url{http://cseweb.ucsd.edu/~lvdmaaten/hucrf/Hidden-Unit_Conditional_Random_Fields.html}}.  

2.  ICDAR 2003 word recognition dataset\footnote{\url{http://algoval.essex.ac.uk/icdar/datasets/TrialTrain/word.zip}}.
After deleting non recognizable numbers and characters, we have 1147 effective words available for training. All of these words belong to 874 classes (words), which consisted of 70 unique characters.

\subsection{Experimental Setup}
In our experiments, we randomly initialized the weight ${\bf W}$ by sampling from the normal Gaussian distribution, and all other parameters in $\boldsymbol{\theta}$ to be zero (i.e. biases $\boldsymbol{b}$ and $ \boldsymbol{c}$, and the transition matrix ${\bf A}$ all to be zero). As for $\boldsymbol{\omega} = \{ {\bf W}_l |  l \in [1,..,L-1] \} $, we initialized them with DBN, which had been mentioned before. As for the number of layers and the number of hidden units in each layer, we set differently according to the dimensionality for different datasets. In all the experiments, we used the 3-layer deep CRFs model with hidden units [400 200 100] respectively in each layer. As for context model, we used the second order potentials over characters to recognize words. 

\begin{table}
\centering
\resizebox{0.7\columnwidth}{!}{%
\begin{tabular}{ |l|c| }
\hline
 \multicolumn{2}{ |c| }{{\bf Word recognition error rate} (\%)} \\\cline{1-2}
 Linear-chain CRF \cite{Lafferty01} & 53.2 \\
\hline
LSTM \cite{Graves08} & 2.31 \\
\hline
DNN \cite{Hinton06b} & 18.5 \\
\hline
Hidden-unit CRF  \cite{Maaten11} &  4.62\\
\hline
Traditional DNN+CRFs \cite{Do10}& 5.19 \\
\hline
Our method  & {\bf 1.60} \\
\hline
\end{tabular}
}
\caption{The experimental comparisons on the OCR dataset. The results reveal the merits of our method, and show that our deep CRFs outperforms other methods significantly.}
\label{tab:ocr}
\end{table}

Evaluation: we use word error rate to measure all methods in 10-fold cross-validation. 
In all experiments with perceptron learning, we did not use regularization terms. In other words, we set $\lambda_2 =0$. And $\lambda_3= 2\times10^{-4}$ for weights in the deep network. For each dataset, we followed the protocol in \cite{Maaten11} and divided it into $10$ folds (9 folds as the training set, and the rest as the testing/validation set), and performed 100 full sweeps through the training data, to update the model parameters. We tuned the base step size based on the error on a small held-out validation set. From the base step size, we computed parameter-specific step sizes $\eta_{\boldsymbol{\theta} }$ and $\eta_{\boldsymbol{\omega} }$ as suggested by \cite{Gelfand10}. %

In Table \ref{tab:ocr}, we compared the performance of our method with the performance of competing models on the OCR dataset. Our method yields a generalization word error of $1.6\%$, while the best performance of other methods is $2.31\%$. It also demonstrates that our model is significantly better than other methods, and the deep structure is definitely helpful than the shadow models, such as hidden CRFs. In addition, our method is significantly better than traditional DNN+CRFs. Thus, as for our model, it shows that the learning stage is helpful and can improve the accuracy significantly. 

We also tested our method on the ICDAR2003 dataset. In our experiment, we first split the words into characters, and then we tested our deep CRFs over the characters. As for word segmentation, we use an naive approach (more sophisticated word segmentation methods will be definitely helpful in this case). Basically, given the word image, we know the number of characters in this image according to its label. Thus, we split the word image equally by dividing the number of characters. Since the sizes of all images are different, we need to resize all characters into the same dimension. Hence, we computed the width-height ratio for each character (which is the split one from its corresponding word image) on all the dataset, and then get the average size for all characters. The result character size is $65\times40$, and we reshape it to a binary vector with 2600 dimension (Note that we binarize all the images or characters). Table \ref{tab:icdar2003} shows the word recognition error rate on ICDAR2003 dataset with five folder cross validation. Again, it demonstrates that our method outperforms CRFs related models and deep learning methods. It also indicates that our model is better than traditional DNN+CRFs. 
\begin{figure}[!t]
\centering
\includegraphics[trim = 50mm 95mm 40mm 95mm, clip, width=8.5cm]{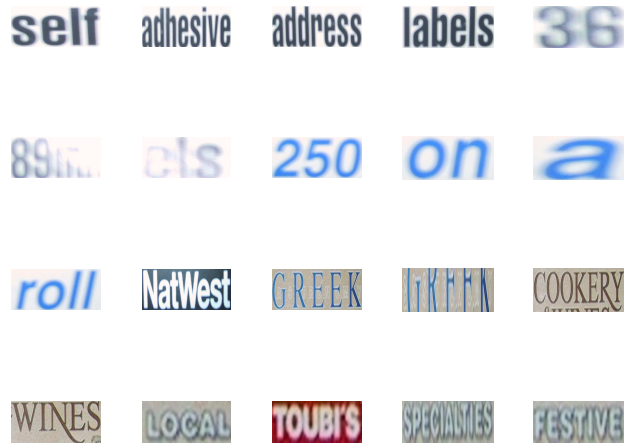}
\caption{The samples from the ICDAR 2003 word recognition dataset. The subset in our experiment has total 70 unique characters and 874 words.}
\label{fig_icdar}
\end{figure}

\begin{table}
\centering
\resizebox{0.7\columnwidth}{!}{%
\begin{tabular}{ |l|c| }
\hline
 \multicolumn{2}{ |c| }{{\bf word recognition (error rate)} (\%)} \\\cline{1-2}
 Linear-chain CRF \cite{Lafferty01} & 97.6 \\
\hline
DNN \cite{Hinton06b} & 41.8 \\
\hline
Hidden-unit CRF  \cite{Maaten11} &  62.7\\
\hline
 Traditional DNN+CRFs \cite{Do10} & 82.6 \\
\hline
Our method & {\bf 40.2} \\
\hline
\end{tabular}
}
\caption{The experimental comparisons on the ICDAR2003 word recognition dataset. The results reveal that our deep CRFs is effective for word recognition.}
\label{tab:icdar2003}
\end{table}

\section{Conclusion}
In this paper, we propose a deep conditional random fields (deep CRFs) for word classification problems. 
Compared to traditional DNN+CRFs, we propose a mixture online learning algorithm: perceptron training for CRFs parameters and stochastic gradient descent for low level weights in the deep structure. We update parameters in an online fashion, which makes it possible to apply our model to large scale datasets. We tested our methods on widely used word recognition datasets, and show that our deep CRFs is effective compared to other shallow CRFs and deep learning methods. 




%

\bibliography{hrbib}
\bibliographystyle{IEEEtran}

\end{document}